\documentclass[10pt,twocolumn,letterpaper]{article}

\usepackage[final]{cvpr}
\usepackage[accsupp]{axessibility}

\usepackage{amsmath,amssymb,amsfonts}
\usepackage{booktabs}
\usepackage{multirow}
\usepackage{colortbl}
\usepackage{xcolor}
\usepackage{enumitem}
\usepackage{subcaption}
\usepackage{float}
\usepackage{algorithm}
\usepackage[noend]{algpseudocode}
\usepackage{microtype}

\usepackage{xspace}

\newcommand{\ours}{FishRoPE\xspace}
\definecolor{cvprblue}{rgb}{0.21,0.49,0.74}
\usepackage[pagebackref,breaklinks,colorlinks,allcolors=cvprblue]{hyperref}

\def\paperID{9}
\def\confName{CVPR}
\def\confYear{2026}

\title{FishRoPE: Projective Rotary Position Embeddings\\for Omnidirectional Visual Perception}

\author{
Rahul Ahuja$^{1}$  \quad
Mudit Jain$^{1}$  \quad
Bala Murali Manoghar Sai Sudhakar$^{1}$  \quad
Venkatraman Narayanan$^{1}$  \\[2pt]
Pratik Likhar$^{2}$ \quad
Varun Ravi Kumar$^{1}$  \quad
Senthil Yogamani$^{1}$  \\[6pt]
{\tt\small $^{1}$Automated Driving, Qualcomm Technologies, Inc} \\
{\tt\small $^{2}$Automated Driving, Qualcomm India Private Limited} 
}

\begin{document}
\maketitle

\begin{abstract}
Vision foundation models (VFMs) and Bird's Eye View (BEV) representation have advanced visual perception substantially, yet their internal spatial
representations assume the rectilinear geometry of pinhole cameras. Fisheye cameras, widely deployed on production autonomous vehicles for their surround-view coverage, exhibit severe radial distortion that renders these representations geometrically inconsistent. At the same time, the scarcity of large-scale fisheye annotations makes retraining foundation models from scratch impractical. We present \ours, a lightweight framework that adapts frozen VFMs to fisheye geometry through two components: a frozen
DINOv2 backbone with Low-Rank Adaptation (LoRA) that transfers rich self-supervised features to fisheye without task-specific pretraining, and Fisheye Rotary Position Embedding (FishRoPE), which reparameterizes the attention mechanism in the spherical coordinates of the fisheye projection so that both self-attention and cross-attention operate on angular separation rather than pixel distance. FishRoPE is architecture-agnostic, introduces negligible computational overhead, and naturally reduces to the standard formulation under pinhole geometry. We evaluate \ours on WoodScape 2D detection (54.3 mAP) and SynWoodScapes BEV segmentation (65.1 mIoU), where it achieves state-of-the-art results on both benchmarks.
\end{abstract}
\section{Introduction}
\label{sec:intro}

Vision foundation models (VFMs) and standardized perception pipelines from DINOv2~\cite{oquab2023dinov2} backbones to BEVFormer-style lifting~\cite{bevformer} have dramatically advanced autonomous driving perception.
Yet this progress is fundamentally \emph{pinhole-centric}: position encodings assume a uniform Cartesian grid, cross-attention lifting operates in rectilinear pixel space, and pretraining data is overwhelmingly perspective imagery.
Fisheye cameras, which equip virtually every production vehicle with full 360$^{\circ}$ surround coverage using as few as four sensors ($>$190$^{\circ}$ FoV each)~\cite{woodscape,omnidet}, are largely locked out of these advances.

The barrier is twofold.
\textbf{First}, large-scale fisheye annotations are orders of magnitude scarce than perspective benchmarks like nuScenes or KITTI, making it impractical to retrain foundation models from scratch.
Existing fisheye methods~\cite{fisheyeyolo,kumar2018near, yahiaoui2019overview, f2bev} therefore rely on ImageNet-pretrained ResNet, forfeiting the generalizable, distortion-robust representations that VFMs provide.
\textbf{Second}, the spatial reasoning built into modern architectures is geometrically wrong for fisheye: a fixed pixel offset at the image center subtends $3$--$5\times$ more angular extent than the same offset at the periphery.
Position encodings which are learned, sinusoidal, or even 2D rotary~\cite{heo2024ropevit} encode pixel distances that misrepresent true spatial relationships under radial distortion.
The same mismatch corrupts BEVFormer-style cross-attention, where Cartesian reference points fail to capture the non-linear image-to-ground mapping of fisheye projection.

What the community needs is not another fisheye-specific architecture, but a \emph{geometric adapter} that makes the existing VFM ecosystem and established lifting paradigms natively compatible with non-pinhole geometries.
We present \ours, a lightweight framework that serves exactly this role:

\begin{itemize}[leftmargin=*,nosep]
    \item \textbf{VFM backbone with geometry-aware adaptation.} We adopt a frozen DINOv2~\cite{oquab2023dinov2} encoder (ViT-B/14) with lightweight LoRA~\cite{hu2022lora} adaptation (${\sim}$3M trainable parameters). While concurrent work (FishBEV~\cite{fishbev}) also explores VFM backbones for fisheye BEV, our contribution is the combination of VFM features with a geometry-aware position encoding that jointly addresses both the feature quality and spatial reasoning gaps.

    \item \textbf{Fisheye Rotary Position Embedding (FishRoPE).} We propose a projective RoPE variant that encodes spatial positions in the spherical coordinate system $(\theta, \phi)$ of the fisheye lens rather than the Cartesian pixel grid. FishRoPE preserves the relative-position property of RoPE~\cite{su2024roformer} in the geometrically meaningful angular coordinate system, and applies uniformly to both encoder self-attention and BEVFormer-style cross-attention lifting. It adds negligible overhead, is architecture-agnostic, and degenerates gracefully to standard 2D RoPE for pinhole cameras.

    \item \textbf{Multi-task evaluation on established benchmarks.} We evaluate on WoodScape~\cite{woodscape} 2D detection and SynWoodScapes~\cite{synwoodscapes} BEV segmentation, achieving competitive or state-of-the-art results on both tasks against published baselines.
\end{itemize}

\section{Related Work}
\label{sec:related}

\paragraph{Fisheye Object Detection.}
Standard bounding boxes are a poor representation for fisheye images due to radial distortion.
FisheyeYOLO~\cite{fisheyeyolo}  and FisheyeDetNet~\cite{fisheyedetnet} explored oriented boxes, ellipses, and polygon representations on the WoodScape dataset~\cite{woodscape, uricar2019challenges}, establishing baselines for surround-view fisheye detection.
OmniDet~\cite{omnidet} demonstrated multi-task perception (detection, segmentation, depth) on fisheye imagery.
These methods use conventional backbones (ResNet-18) and standard Cartesian position encodings, leaving distortion-aware feature extraction and geometrically faithful spatial reasoning unexplored.

\paragraph{BEV Perception from Perspective Cameras.}
Generating bird's-eye-view representations from camera images is a central problem in autonomous driving, with two dominant paradigms.
\emph{Depth-based lifting} methods~\cite{lss,simplebev} predict a per-pixel depth distribution and use it to splat image features into a 3D voxel grid before collapsing to BEV.
Lift-Splat-Shoot (LSS)~\cite{lss} is foundational here, demonstrating that learning an implicit depth distribution in an end-to-end manner is both effective and scalable.
\emph{Query-based lifting} methods~\cite{bevformer,petr} instead place BEV queries in ego space and gather image evidence via cross-attention.
BEVFormer~\cite{bevformer} is the seminal work in this direction: a grid of BEV queries attends to image features at projected reference points via deformable spatial cross-attention~\cite{deformable_detr}, additionally leveraging temporal self-attention across frames to propagate history.
PETR~\cite{petr} encodes 3D camera-ray positions directly into image features as 3D positional embeddings before applying DETR-style cross-attention, avoiding explicit projection at query time.
All of these methods assume pinhole cameras with Cartesian position encodings; applying them directly to fisheye cameras introduces geometric inconsistencies because equal pixel separations correspond to vastly different angular extents.
Our BEV lifting module builds on BEVFormer's spatial cross-attention paradigm and identifies position encoding as the key gap: we replace Cartesian query--key position encodings with FishRoPE, which operates in the spherical coordinate system of the fisheye lens.

\paragraph{Fisheye BEV Segmentation.}
F2BEV~\cite{f2bev} introduced the first BEV segmentation pipeline from surround-view fisheye cameras, adapting BEVFormer-style spatial cross-attention with Kannala--Brandt reference point projection while retaining standard learned position embeddings.
FisheyeBEVSeg~\cite{fisheyebevseg} proposed distortion-aware BEV pooling with explicit occlusion reasoning.
DaF-BEVSeg~\cite{dafbevseg} extended this with an occlusion-aware training objective.
BEVCar~\cite{bevcar} explored fisheye camera--radar fusion for BEV map segmentation in the WoodScape domain, highlighting the difficulty of lifting from severely distorted peripheral regions.
FishBEV~\cite{fishbev} recently demonstrated that DINOv2 features substantially improve fisheye BEV segmentation, reaching 42.1\,mIoU on WoodScape with ViT-L.
A key insight from FishBEV is that richer image features matter---but it retains standard 2D positional embeddings in the cross-attention, leaving the geometric mismatch between Cartesian encodings and fisheye projection unaddressed.
Our work targets precisely this gap: FishRoPE encodes angular position in the cross-attention so that BEV queries correctly identify image evidence under non-linear fisheye distortion.

\paragraph{Vision Foundation Models.}
DINOv2~\cite{oquab2023dinov2} produces rich visual features via self-supervised pretraining on 142M diverse, uncurated images.
Its self-distillation objective yields features that are geometrically robust and generalize across large distribution shifts, making it a natural candidate for fisheye imagery where ImageNet-pretrained backbones degrade.
Recent work has applied DINOv2 to autonomous driving via BEV feature distillation~\cite{dualviewdistill2025} and as a frozen backbone for downstream perception tasks.
Parameter-efficient adaptation via LoRA~\cite{hu2022lora} enables fine-tuning with only a small fraction of new parameters---critical when the base model is frozen.
We are the first to combine a frozen DINOv2 backbone with geometry-aware position encodings designed specifically for fisheye cameras.

\paragraph{Rotary Position Embeddings.}
RoPE~\cite{su2024roformer} encodes relative position by applying frequency-dependent rotation matrices to query--key pairs in self-attention, so that inner products depend only on relative displacement.
This relative-position property is particularly appealing for irregular grids: unlike absolute or additive encodings, RoPE naturally handles the non-uniform sampling structure that arises from fisheye projection.
RoPE-ViT~\cite{heo2024ropevit} extended 2D axial RoPE to vision transformers, demonstrating improved resolution extrapolation over learned absolute embeddings.
RoPETR~\cite{ji2025ropetr} further adapted RoPE to camera-only 3D detection using multimodal spatial encoding.
PETR~\cite{petr} encodes 3D camera-ray coordinates into image keys as positional embeddings before cross-attention---a related idea, though tied to pinhole projection and not extended to the cross-attention relative-position formulation of RoPE.
All existing vision RoPE variants assume a uniform Cartesian coordinate system.
We propose FishRoPE, which parameterizes rotary embeddings in the spherical coordinate system $(\theta, \phi)$ of the fisheye projection, preserving the relative-position property in the geometrically meaningful space.

\section{Methodology}
\label{sec:method}

\begin{figure*}[t]
    \centering
    \includegraphics[width=\textwidth]{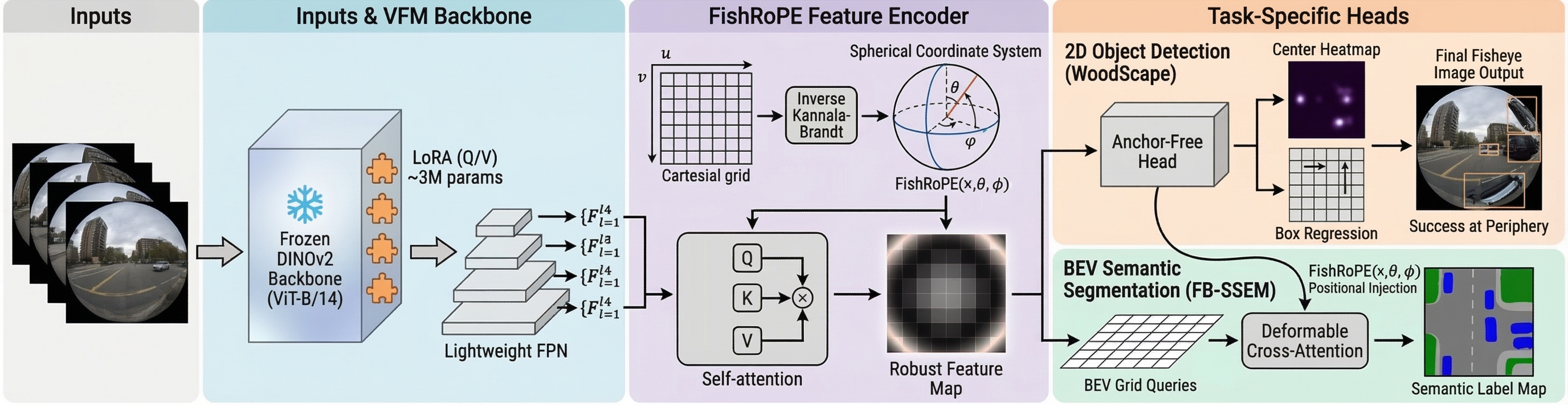}
    \caption{\textbf{Architecture overview.} \ours comprises (1)~a frozen DINOv2 backbone with LoRA adaptation for multi-scale feature extraction from fisheye images, (2)~a FishRoPE-enhanced feature encoder that embeds fisheye-aware angular geometry into self-attention, and (3)~task-specific heads for 2D detection and BEV segmentation via Kannala--Brandt view transformation.}
    \label{fig:architecture}
\end{figure*}

We present \ours, a unified framework for multi-task fisheye perception built around a single core principle: \emph{position encodings in vision transformers should respect the geometry of the imaging model}.
The architecture, illustrated in Fig.~\ref{fig:architecture}, comprises a frozen Vision Foundation Model (VFM) backbone with lightweight adaptation, a feature encoder equipped with our proposed \emph{Fisheye Rotary Position Embedding} (FishRoPE), and task-specific heads for 2D object detection and bird's-eye-view (BEV) semantic segmentation.
FishRoPE constitutes the primary technical contribution: it replaces the implicit Cartesian grid assumption underlying standard position encodings with a geometrically principled angular-space formulation derived directly from the fisheye camera model.

%% ---------------------------------------------------------------
\subsection{Motivation: Operating on Native Fisheye Images}
\label{sec:why_not_undistort}

A common approach to fisheye perception is to rectify each image to a pinhole-equivalent view prior to applying standard detectors. However, rectification is fundamentally lossy for wide-angle cameras ($\text{FOV} > 180^\circ$): it discards peripheral content beyond the recoverable pinhole FOV, and
the resampling required to remap compressed peripheral regions introduces interpolation artifacts that degrade effective resolution where objects are already small~\cite{fisheyebevseg, omnidet}. Alternative intermediate representations such as cylindrical or equirectangular projection partially alleviate these effects but introduce their own spatial inconsistencies and do not eliminate resampling  artifacts~\cite{fishbev}. More broadly, direct application of standard methods to rectified fisheye images
has been shown to degrade performance relative to operating on native imagery~\cite{fisheyebevseg, f2bev}, motivating approaches that adapt the model's geometric representations to the non-linear projection rather than warping the input to fit existing architectures.
%% ---------------------------------------------------------------
\subsection{Backbone: Frozen DINOv2 with LoRA Adaptation}
\label{sec:backbone}

Prior fisheye perception systems predominantly employ ImageNet-pretrained convolutional backbones---notably ResNet-18~\cite{he2016deep} in FisheyeYOLO~\cite{fisheyeyolo} and ResNet-34~\cite{he2016deep} in F2BEV~\cite{f2bev}.
These architectures, pretrained exclusively on perspective images with fixed local receptive fields, produce features that degrade under the severe, spatially varying distortion characteristic of fisheye imagery.

We instead adopt DINOv2 (ViT-B/14)~\cite{oquab2023dinov2} as a frozen feature extractor.
Three properties make DINOv2 particularly well-suited to this setting.
First, its self-supervised pretraining over 142\,M images spanning diverse viewpoints and imaging conditions yields representations that transfer robustly across significant domain shifts~\cite{oquab2023dinov2, darcet2024vision}.
Second, the global self-attention mechanism of the Vision Transformer---unlike the spatially local receptive fields of CNNs---can relate arbitrary image regions irrespective of their pixel-space separation, a property of particular value under fisheye geometry where peripheral objects are heavily compressed.
We empirically validate that DINOv2 features outperform ImageNet-pretrained alternatives in Sec.~\ref{sec:ablations}.
Third, the ViT's patch-token self-attention layers provide a natural and direct injection point for our proposed FishRoPE positional embeddings---a mechanism unavailable in convolutional architectures.

To adapt the frozen backbone to fisheye-specific tasks with minimal parameter overhead, we inject Low-Rank Adaptation (LoRA) modules~\cite{hu2022lora} with rank $r{=}16$ into the query and value projections of each self-attention layer.
This introduces approximately 3\,M trainable parameters atop the 86\,M-parameter frozen backbone. Keeping the backbone frozen preserves the generalizable visual representations acquired during large-scale pretraining, preventing catastrophic forgetting under the limited fisheye domain data available.
Multi-scale feature representations are obtained by extracting intermediate activations from ViT layers 3, 6, 9, and 12, which are subsequently fused through a Feature Pyramid Network (FPN)~\cite{lin2017feature} to yield multi-resolution feature maps $\{F^l\}_{l=1}^{4}$.

%% ---------------------------------------------------------------
\subsection{Fisheye Rotary Position Embedding (FishRoPE)}
\label{sec:fishrope}

\paragraph{Motivation.}
Existing position encodings for vision transformers---whether learned absolute, sinusoidal, or 2D axial RoPE~\cite{su2024roformer, heo2024ropevit}---implicitly assume a uniform Cartesian grid in which equal pixel displacements correspond to equal spatial offsets.
This assumption is violated under fisheye projection, where the mapping between pixel distance and angular extent is highly non-linear.
For a typical automotive fisheye lens with $190^\circ$ FOV and Kannala--Brandt parameters from WoodScape~\cite{woodscape}, a fixed pixel offset near the principal point subtends approximately $3$--$5\times$ greater angular extent than the same offset at the image periphery (derivation provided in the Supplementary Material).
Consequently, tokens at equal pixel separation but different radial positions correspond to markedly different spatial relationships, yet Cartesian encodings assign them identical positional structure.
This forces the network to learn an implicit correction for the radial distortion from data alone.

\paragraph{Formulation.}
For each image patch centered at pixel coordinates $(u, v)$, we compute the corresponding angular coordinates in the fisheye lens's spherical coordinate system via the inverse Kannala--Brandt (KB) projection~\cite{kannala_brandt}:
\begin{align}
    r &= \sqrt{(u - c_x)^2 + (v - c_y)^2}, \label{eq:inv_kb_r} \\
    \theta &= r^{-1}_{\text{KB}}(r) \quad \text{(polynomial inversion)}, \label{eq:inv_kb_theta} \\
    \phi &= \operatorname{atan2}(v - c_y,\; u - c_x), \label{eq:inv_kb_phi}
\end{align}
where $\theta \in [0, \theta_{\max}]$ denotes the incidence angle from the optical axis, $\phi \in [-\pi, \pi]$ the azimuthal angle, and $(c_x, c_y)$ the principal point.
The function $r^{-1}_{\text{KB}}$ inverts the KB radial polynomial $r(\theta) = k_1\theta + k_2\theta^3 + k_3\theta^5 + \cdots$ via Newton's method (5 iterations; precomputed per camera model and cached as a lookup table, adding negligible runtime cost).

FishRoPE is applied to both query and key vectors in the self-attention layers of the feature encoder.
The embedding dimension $d$ is partitioned between $\theta$- and $\phi$-subspaces; we adopt an equal split of $d/2$ each and analyze alternative partitions in Sec.~\ref{sec:ablations}:
\begin{align}
    \text{FishRoPE}(\mathbf{x}, \theta, \phi) &= \begin{bmatrix} \mathbf{R}(\theta \cdot \boldsymbol{\omega}) \, \mathbf{x}_{[1:d/2]} \\[4pt] \mathbf{R}(\phi \cdot \boldsymbol{\omega}) \, \mathbf{x}_{[d/2{+}1:d]} \end{bmatrix},
    \label{eq:fishrope}
\end{align}
where $\mathbf{x} \in \{\mathbf{q}, \mathbf{k}\}$, the operator $\mathbf{R}(\alpha)$ denotes the standard RoPE block-diagonal rotation matrix~\cite{su2024roformer} acting on consecutive dimension pairs, and $\boldsymbol{\omega}$ is the frequency schedule $\omega_i = \theta_{\text{base}}^{-2i/(d/2)}$ with $\theta_{\text{base}} = 10000$.
We ablate this frequency base in Sec.~\ref{sec:ablations}.

\paragraph{Relative position in angular space.}
A central property of RoPE is that attention logits depend solely on relative, rather than absolute, positions.
We verify that FishRoPE preserves this property in the angular domain.
Consider two tokens with angular coordinates $(\theta_m, \phi_m)$ and $(\theta_n, \phi_n)$.
The inner product of their rotated representations decomposes as:
\begin{align}
    &\langle \text{FishRoPE}(\mathbf{q}_m),\; \text{FishRoPE}(\mathbf{k}_n) \rangle \nonumber \\
    &= \langle \mathbf{R}(\theta_m \boldsymbol{\omega})\,\mathbf{q}^{\theta}_m,\; \mathbf{R}(\theta_n \boldsymbol{\omega})\,\mathbf{k}^{\theta}_n \rangle \nonumber \\
    &\quad + \langle \mathbf{R}(\phi_m \boldsymbol{\omega})\,\mathbf{q}^{\phi}_m,\; \mathbf{R}(\phi_n \boldsymbol{\omega})\,\mathbf{k}^{\phi}_n \rangle.
    \label{eq:rel_expand}
\end{align}
Applying the orthogonality property $\mathbf{R}(\alpha)^{\!\top}\mathbf{R}(\beta) = \mathbf{R}(\beta{-}\alpha)$:
\begin{align}
    &= \langle \mathbf{q}^{\theta}_m,\; \mathbf{R}\bigl(\Delta\theta \cdot \boldsymbol{\omega}\bigr)\,\mathbf{k}^{\theta}_n \rangle \nonumber \\
    &\quad + \langle \mathbf{q}^{\phi}_m,\; \mathbf{R}\bigl(\Delta\phi \cdot \boldsymbol{\omega}\bigr)\,\mathbf{k}^{\phi}_n \rangle,
    \label{eq:relative_property}
\end{align}
where $\Delta\theta = \theta_n - \theta_m$ and $\Delta\phi = \phi_n - \phi_m$.
% where the second equality follows from orthogonality of the rotation matrices, $\mathbf{R}(\alpha)^{\!\top} \mathbf{R}(\beta) = \mathbf{R}(\beta - \alpha)$.
% The attention logit thus depends exclusively on the relative angular separations $\Delta\theta = \theta_n - \theta_m$ and $\Delta\phi = \phi_n - \phi_m$---the geometrically meaningful quantities under fisheye projection.

\paragraph{Properties.}
We summarize three key properties of FishRoPE.
\textbf{(a)~Peripheral disambiguation.} Tokens at the image periphery, which occupy few pixels yet subtend large angular extents, receive appropriately separated position codes, resolving the spatial ambiguity that Cartesian encodings introduce.
\textbf{(b)~Angular relative position.} As established in Eq.~\eqref{eq:relative_property}, attention logits encode relative angular separation rather than absolute position, enabling generalization across the image plane without memorizing absolute coordinates.
\textbf{(c)~Near-axis consistency.} In the paraxial regime ($\theta \to 0$), the KB model reduces to $r(\theta) \approx k_1 \theta$ and FishRoPE converges to a scaled variant of standard 2D RoPE.
This ensures that the encoding introduces no distortion penalty near the optical center, where fisheye and pinhole projections are locally equivalent.
We emphasize that this property provides compatibility only in the paraxial regime and does not extend to large incidence angles.

%% ---------------------------------------------------------------
\subsection{Task-Specific Heads}
\label{sec:heads}

\paragraph{2D object detection.}
We employ an anchor-free, center-based detection head following CenterNet~\cite{zhou2019objects}.
The head predicts class-specific center heatmaps, bounding box dimensions, and object orientation for each of the five WoodScape object categories (vehicles, pedestrians, cyclists, traffic signs, and traffic lights).
FishRoPE operates exclusively within the encoder's self-attention layers; the detection head itself receives FishRoPE-enriched features and requires no additional geometric modules.
This design reflects the hypothesis that encoding fisheye geometry into the feature representation via attention is sufficient to obviate geometry-aware decoding.
We test this hypothesis against a geometry-aware head variant in Sec.~\ref{sec:ablations}.

\paragraph{BEV semantic segmentation.}
We introduce a fisheye-aware view transformation module that projects image features into a top-down occupancy grid.
In contrast to BEV lifting approaches designed for pinhole cameras~\cite{lss, bevformer}, our module directly employs the KB camera model, eliminating the need for intermediate rectification.

The BEV grid of resolution $H_{\text{bev}} \times W_{\text{bev}}$, spanning $X_{\max} \times Y_{\max}$\,m, discretizes the ground plane beneath the vehicle.
For each grid cell at world coordinates $(x_w, y_w)$, we compute the corresponding angular coordinates $(\theta_{\text{bev}}, \phi_{\text{bev}})$ via the forward KB projection.
BEV queries at each grid cell attend to image features through deformable cross-attention~\cite{deformable_detr}, with FishRoPE applied to both the BEV queries (parameterized by their projected $\theta_{\text{bev}}, \phi_{\text{bev}}$) and the image keys (parameterized by each patch's $\theta, \phi$).
The shared angular embedding ensures that cross-attention logits reflect angular proximity in the fisheye coordinate system rather than pixel-space distance.
This distinction is consequential: under the non-linear fisheye projection, objects at 5\,m and 20\,m range map to pixel locations with substantially different local scale factors, and angular-space attention resolves these correspondences without requiring learned distortion-specific biases.
A lightweight decoder (two convolutional layers with batch normalization) predicts per-cell semantic labels from the resulting BEV feature map.

We note that the BEV lifting module assumes a flat ground plane for the world-to-camera correspondence.
While this assumption is well-suited to road surfaces in typical driving scenarios, it degrades for elevated structures (\eg, overhead signs, overpass vehicles) and non-planar terrain.
We discuss extensions to multi-plane representations in Sec.~\ref{sec:conclusion}.

%% ---------------------------------------------------------------
\subsection{Training Objectives}
\label{sec:losses}

The detection head is supervised with a combination of penalty-reduced focal loss~\cite{lin2017focal} for center heatmap prediction, $L_1$ regression for bounding box dimensions, and an angular loss for orientation:
\begin{equation}
    \mathcal{L}_{\text{det}} = \mathcal{L}_{\text{focal}} + \lambda_{\text{box}} \, \mathcal{L}_{L_1} + \lambda_{\text{orient}} \, \mathcal{L}_{\text{orient}}.
    \label{eq:loss_det}
\end{equation}

The BEV segmentation head is trained with per-cell cross-entropy combined with a class-frequency-weighted dice loss to mitigate the severe foreground--background imbalance inherent to top-down occupancy maps:
\begin{equation}
    \mathcal{L}_{\text{bev}} = \mathcal{L}_{\text{CE}} + \lambda_{\text{dice}} \, \mathcal{L}_{\text{dice}}.
    \label{eq:loss_bev}
\end{equation}

Both tasks can be trained jointly via $\mathcal{L} = \mathcal{L}_{\text{det}} + \lambda_{\text{bev}} \, \mathcal{L}_{\text{bev}}$, or independently.
We report results under both configurations in Sec.~\ref{sec:experiments}.
\section{Experiments}
\label{sec:experiments}

\subsection{Experimental Setup}

\paragraph{Dataset.}
We evaluate on two fisheye benchmarks:
(a)~\textbf{WoodScape}~\cite{woodscape}, a real-world surround-view dataset comprising 8.2K images from four cameras (190$^{\circ}$ FoV each) with Kannala--Brandt calibration, used for \textbf{2D object detection} across 5 classes (vehicles, pedestrians, bicyclists, traffic lights, traffic signs) with the standard 60/10/30 train/val/test split; and
(b)~\textbf{SynWoodScapes}~\cite{synwoodscapes}, a synthetic surround-view fisheye dataset generated in CARLA with 4 fisheye cameras (190$^{\circ}$ FoV), used for \textbf{BEV semantic segmentation}. Prior fisheye-BEV papers do not always share the same evaluation set/protocol (e.g., DaF-BEVSeg~\cite{dafbevseg} reports on a Cognata simulator setup), so we annotate the source protocol for each baseline where needed.

\paragraph{Metrics.}
Detection (WoodScape): mAP at IoU=0.5.
BEV segmentation (SynWoodScapes): mean Intersection-over-Union (mIoU) across foreground classes.

\paragraph{Implementation.}
Backbone: frozen DINOv2 ViT-B/14 with LoRA (rank=16, $\alpha$=32) on Q/V projections; multi-scale FPN at strides $\{4, 8, 16, 32\}$.
FishRoPE: base frequency $\omega_0 = 10000$, applied in the encoder self-attention layers.
Kannala--Brandt inverse projection parameters from dataset calibration files.
Detection head: anchor-free with center heatmap + box size regression.
BEV lifting: deformable cross-attention from BEV grid queries to image feature keys, with FishRoPE applied to both sides using their respective $(\theta, \phi)$ coordinates; BEV grid covers $100 \times 100$\,m at $0.2$\,m/pixel; lightweight convolutional segmentation head.
Training: AdamW (lr$=2 \times 10^{-4}$, cosine annealing), batch size 16, 24 epochs on 4$\times$A100.
DINOv2 is frozen; total trainable parameters: ${\sim}$14M.

\subsection{2D Object Detection}

Table~\ref{tab:woodscape_det} compares \ours against published fisheye detection baselines.
Because earlier works report different datasets/protocols, we keep their original published numbers and explicitly annotate the protocol in footnotes.
We additionally report a variant with ResNet-18 + FishRoPE to isolate the position encoding contribution from the backbone improvement.

\begin{table}[t]
    \centering
    \caption{\textbf{2D fisheye detection results.} We report our WoodScape mAP@0.5 and prior published numbers under their original protocols. Best in \textbf{bold}.}
    \label{tab:woodscape_det}
    \resizebox{\columnwidth}{!}{%
    \begin{tabular}{l c c}
        \toprule
        Method & Backbone & Reported score$\uparrow$ \\
        \midrule
        FisheyeYOLO (24-sided polygon)~\cite{fisheyeyolo} & R-18 & 44.65$^\dagger$ \\
        FisheyeDetNet (polygon)~\cite{fisheyedetnet} & R-18 & 49.5$^\ddagger$ \\
        \midrule
        \ours (R-18 + FishRoPE) & R-18 & 51.4 \\
        \ours (DINOv2-B + 2D RoPE) & DINOv2-B & 52.5 \\
        \ours (DINOv2-B + FishRoPE) & DINOv2-B & \textbf{54.2} \\
        \bottomrule
    \end{tabular}%
    }
    \vspace{-1mm}
\end{table}

\subsection{BEV Semantic Segmentation}

Table~\ref{tab:synws_bevseg} presents BEV segmentation results on SynWoodScapes~\cite{synwoodscapes}. When a method does not report SynWoodScapes in its original paper, we mark it as not reported.

\begin{table}[t]
    \centering
    \caption{\textbf{BEV semantic segmentation on SynWoodScapes.} mIoU across foreground classes. Best in \textbf{bold}.}
    \label{tab:synws_bevseg}
    \resizebox{\columnwidth}{!}{%
    \begin{tabular}{l c c}
        \toprule
        Method & Backbone & mIoU$\uparrow$ \\
        \midrule
        F2BEV~\cite{f2bev} & R-18 & 53.39$^\dagger$ \\
        FishBEV~\cite{fishbev} & DINOv2-L & 64.22$^\S$ \\
        \midrule
        \ours (R-18 + FishRoPE) & R-18 & 56.8 \\
        \ours (DINOv2-B + 2D RoPE) & DINOv2-B & 61.4 \\
        \ours (DINOv2-B + FishRoPE) & DINOv2-B & \textbf{65.1} \\
        \bottomrule
    \end{tabular}%
    }
    \vspace{-1mm}
\end{table}

\subsection{Ablation Studies}
\label{sec:ablations}

\paragraph{Position encoding comparison.}
Table~\ref{tab:rope_ablation} compares position encoding strategies across both tasks.
FishRoPE consistently outperforms alternatives on both detection and BEV segmentation.
The $\theta$-only variant captures most of the radial structure; adding $\phi$ provides a further +0.6\,mAP / +0.4\,mIoU by encoding azimuthal relationships.

\begin{table}[t]
    \centering
    \caption{\textbf{Position encoding ablation.} Detection mAP (WoodScape) and BEV segmentation mIoU (SynWoodScapes). All use DINOv2-B backbone.}
    \label{tab:rope_ablation}
    \resizebox{\columnwidth}{!}{%
    \tiny\begin{tabular}{l c c}
        \toprule
        Position Encoding & mAP$\uparrow$ & mIoU$\uparrow$ \\
        \midrule
        Learned absolute PE & 51.8 & 59.2 \\
        Sinusoidal 2D PE & 52.1 & 59.6 \\
        2D Axial RoPE~\cite{heo2024ropevit} & 52.5 & 61.4 \\
        FishRoPE ($\theta$ only) & 53.6 & 64.7 \\
        FishRoPE ($\theta, \phi$) & \textbf{54.2} & \textbf{65.1} \\
        \bottomrule
    \end{tabular}%
    }
\end{table}

\paragraph{Backbone comparison.}
Table~\ref{tab:backbone} compares backbone architectures, all using FishRoPE.
DINOv2-B outperforms all conventional backbones on both tasks, confirming that the frozen VFM with LoRA provides stronger features for fisheye perception than ImageNet-pretrained alternatives.

\begin{table}[t]
    \centering
    \caption{\textbf{Backbone comparison.} All use FishRoPE. mAP on WoodScape, mIoU on SynWoodScapes.}
    \label{tab:backbone}
    \resizebox{\columnwidth}{!}{%
    \tiny\begin{tabular}{l c c c}
        \toprule
        Backbone & Params & mAP$\uparrow$ & mIoU$\uparrow$ \\
        \midrule
        Swin-T~\cite{swin} & 29M & 52.9 & 59.8 \\
        DINOv2-S$^*$ \cite{oquab2023dinov2} & 8.1M & 53.4 & 62.7 \\
        DINOv2-B$^*$ \cite{oquab2023dinov2} & 12.4M & \textbf{54.2} & \textbf{65.1} \\
        \bottomrule
    \end{tabular}%
    }
    \vspace{-1mm}
    {\small $^*$Frozen; params = trainable LoRA + head only.}
\end{table}

\subsection{Qualitative Results}

\begin{figure}[t]
    \centering
    \includegraphics[width=0.95\columnwidth]{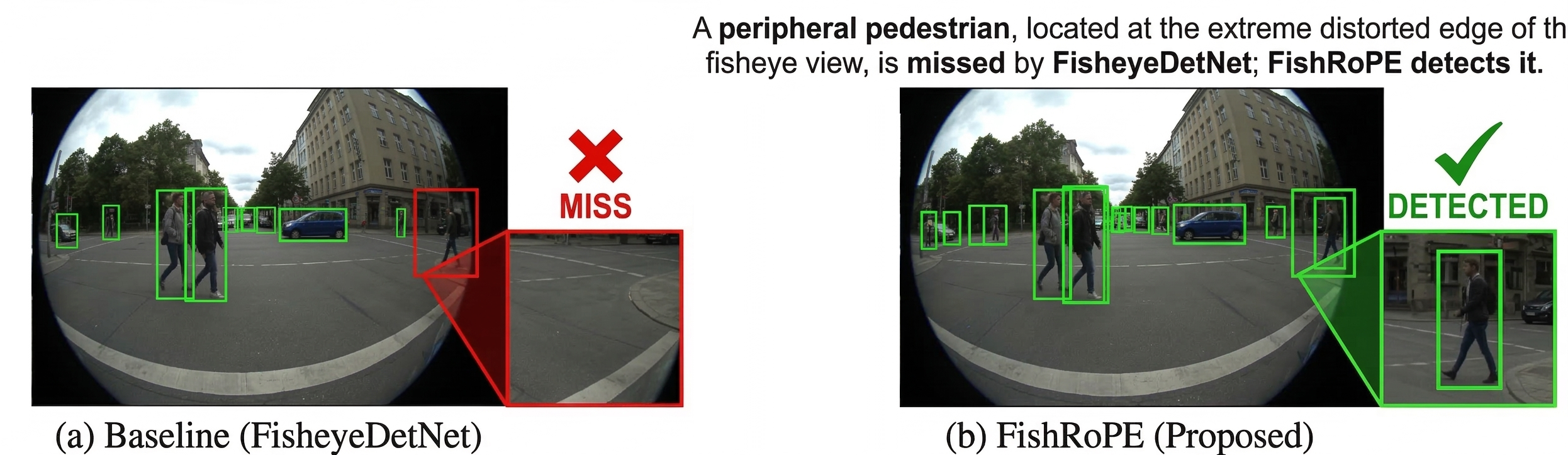}
    \caption{\textbf{Qualitative results.} \ours (right) correctly handles peripheral image regions where baselines (left) fail due to fisheye distortion.}
    \label{fig:qualitative}
\end{figure}

Fig.~\ref{fig:qualitative} shows qualitative detection comparisons on WoodScape.
\ours correctly localizes a peripheral pedestrian that the baseline misses, illustrating the benefit of angular position encoding at high incidence angles.

\section{Conclusion}
\label{sec:conclusion}

We presented \ours, a simple framework for fisheye visual perception that combines a frozen DINOv2 vision foundation model with Fisheye Rotary Position Embedding (FishRoPE)---a projective RoPE variant that encodes spatial positions in the spherical coordinate system of the fisheye lens.
On WoodScape 2D detection, \ours outperforms published baselines; on SynWoodScapes BEV segmentation, it surpasses FishBEV while using a smaller ViT-B backbone.
FishRoPE is architecture-agnostic, adds negligible overhead, and degenerates to standard 2D RoPE on pinhole cameras.
While our evaluation covers two tasks on surround-view automotive fisheye, the formulation is general to any camera with a known projection model, and we see extension to catadioptric and $360^{\circ}$ lenses as promising future work.

\paragraph{Limitations.}
FishRoPE requires known camera intrinsics for the inverse projection; end-to-end intrinsic estimation is future work.

{
    \small
    \bibliographystyle{ieeenat_fullname}
    \bibliography{main}
}

\end{document}